\documentclass[10pt,twocolumn,letterpaper]{article}

\usepackage{cvpr}
\usepackage{times}
\usepackage{epsfig}
\usepackage{graphicx}
\usepackage{mathtools,amsmath}
\usepackage{amssymb}

\usepackage{authblk}
\usepackage{comment}
\usepackage{multirow}
\usepackage{float}
\usepackage[caption = false]{subfig}
\usepackage{adjustbox}
\usepackage[none]{hyphenat}
\usepackage{array}
\newcolumntype{P}[1]{>{\centering\arraybackslash}p{#1}}

\usepackage[pagebackref=true,breaklinks=true,letterpaper=true,colorlinks,bookmarks=false]{hyperref}

\cvprfinalcopy 

\def\cvprPaperID{7121} 

\ifcvprfinal\fi
\begin{document}

\title{Putting visual object recognition in context}

\author[1,2]{Mengmi Zhang}
\author[3]{Claire Tseng}
\author[1,2]{Gabriel Kreiman}

\affil[ ]{\small \{mengmi.zhang@childrens, ctseng@college, gabriel.kreiman@tch\}.harvard.edu\normalsize}
\affil[1]{Children's Hospital, Harvard Medical School}
\affil[2]{Center for Brains, Minds and Machines}
\affil[3]{Harvard College, Harvard University}

\maketitle

\begin{abstract}
Context plays an important role in visual recognition. Recent studies have shown that visual recognition networks can be fooled by placing objects in inconsistent contexts (e.g., a cow in the ocean). To model the role of contextual information in visual recognition, we systematically investigated ten critical properties of where, when, and how context modulates recognition, including the amount of context, context and object resolution, geometrical structure of context, context congruence, and temporal dynamics of contextual modulation. The tasks involved recognizing a target object surrounded with context in a natural image. As an essential benchmark, we conducted a series of psychophysics experiments where we altered one aspect of context at a time, and quantified recognition accuracy. We propose a biologically-inspired context-aware object recognition model consisting of a two-stream architecture. The model processes visual information at the fovea and periphery in parallel, dynamically incorporates object and contextual information, and sequentially reasons about the class label for the target object. Across a wide range of behavioral tasks, the model approximates human level performance \emph{without retraining for each task}, captures the dependence of context enhancement on image properties, and provides initial steps towards integrating scene and object information for visual recognition. All source code and data are publicly available\footnote{\url{https://github.com/kreimanlab/Put-In-Context}}.
\end{abstract}



\vspace{-4mm}
\section{Introduction}

\begin{figure}[t]
\begin{center}
\includegraphics[width=8.3cm]{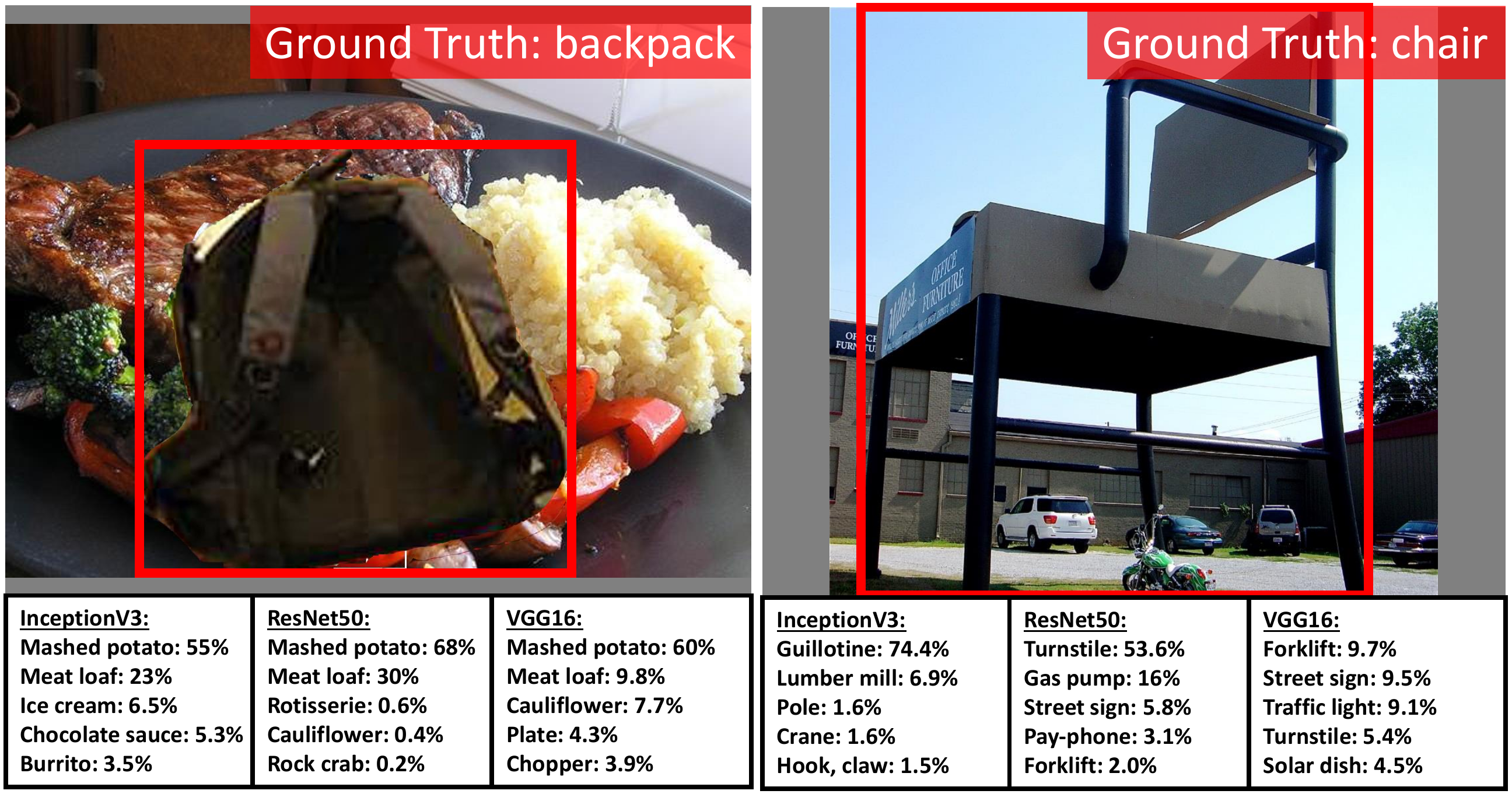}\vspace{-4mm}
\end{center}
   \caption{\textbf{Misclassification of objects in unfamiliar contexts}. State-of-the-art deep visual recognition networks, such as InceptionV3~\cite{szegedy2017inception}, ResNet50~\cite{zagoruyko2016wide}, and VGG16~\cite{simonyan2014very}, make mistakes when the context is incongruent. The top-5 labels and confidence levels by each model are shown at the bottom.
   }\vspace{-5mm}
\label{fig:introincongrueg}
\end{figure}

The tiny object on the table is probably a spoon, not an elephant. Objects do not appear in isolation. Instead, objects co-vary with other objects and scene properties, their sizes and colors usually respect regularities relative to nearby elements, and objects tend to appear at stereotypical locations. The success in object recognition and detection tasks in natural images relies on \emph{implicit} incorporation of contextual information. Deep convolutional neural networks jointly learn  statistical associations between objects, image properties, and  labels ~\cite{choi2012context,sun2017seeing,dvornik2018modeling,beery2018recognition}. Such algorithms can be tricked into mislabeling or missing an object by placing it in an unfamiliar context (Fig.~\ref{fig:introincongrueg}).

\begin{figure*}
\captionsetup[subfloat]{labelformat=empty,position=bottom}
\centering
\subfloat[]{\includegraphics[width= 10.5cm]{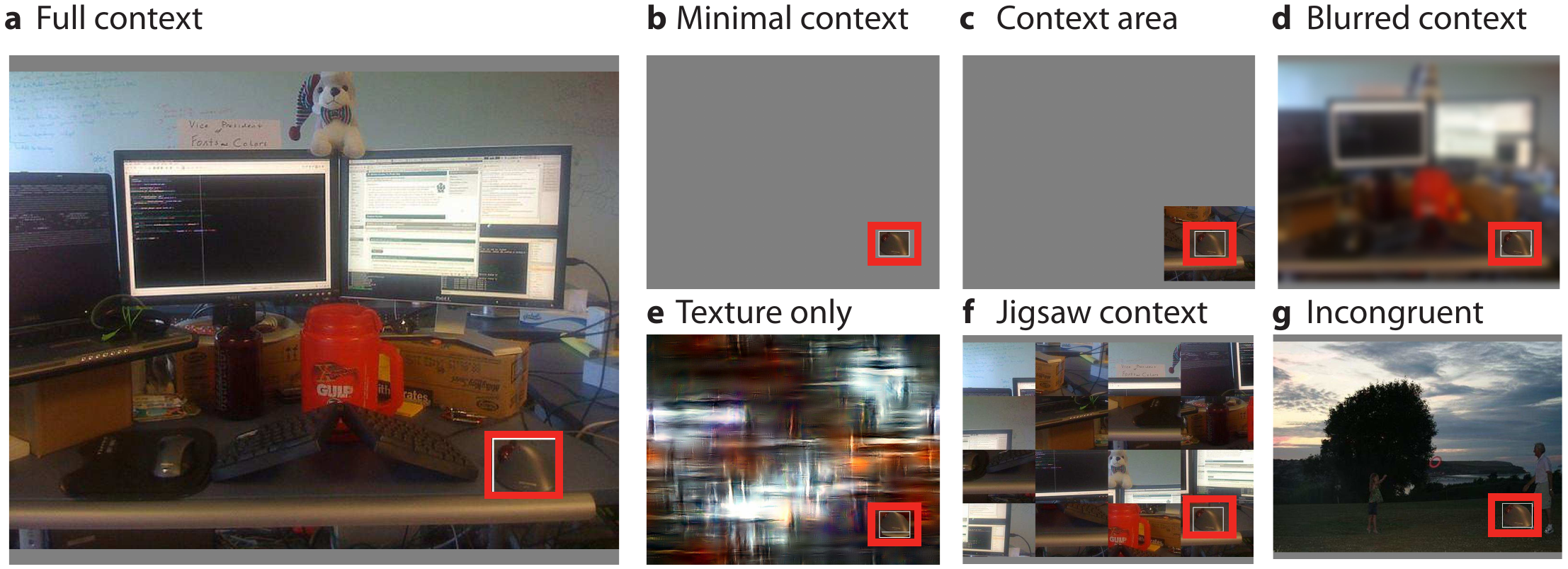}\vspace{-4mm}\label{fig:intro2allcontextconds}}\hspace{0.1cm}
   \subfloat[]{\includegraphics[width= 6.0cm]{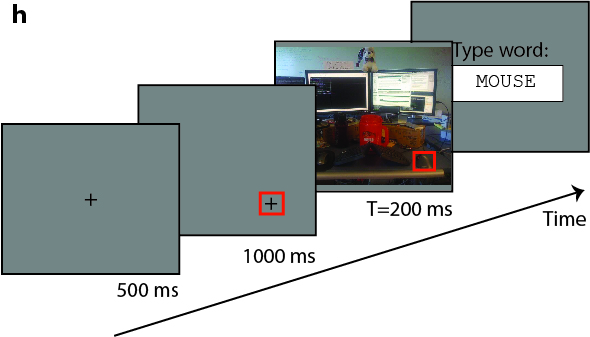}\vspace{-4mm}\label{fig:expschematics}}\vspace{-4mm}
   \caption{\textbf{Fundamental properties of context and task schematic}. Example image with full context (\textbf{a}) and image modifications used in experiments (more examples in Fig. S1). The target location (red box) is always the same across conditions. The correct answer (``mouse") is not shown in the actual experiment). (h) Subjects were presented with a fixation cross (500 ms), followed by a bounding box indicating the target object location (1000 ms). In most experiments (except for Exp C1-3, Fig. S7, S9 and S11), the image  was shown for $T=200$ ms. After image offset, subjects typed one word to identify the target object.}
\label{fig:expsintroall}\vspace{-4mm}
\end{figure*}

Here we systematically and quantitatively investigated the mechanisms by which contextual information is integrated into visual recognition. We focus on three fundamental aspects of context:  [A] the interaction between object size and the amount of contextual information; [B] the geometry, resolution, and content of contextual information; [C] the temporal dynamics of contextual modulation, and the interaction between bottom-up and recurrent computations during contextual modulation. By systematically measuring the effect of context in 10 human psychophysics experiments (Fig.~\ref{fig:expsintroall}, Fig. S1, S7, S9 and S11), we gain a quantitative understanding of where, when, and how context modulates recognition. Moreover, the human data provides a quantitative benchmark to test (but not train)  computational models.

Inspired by the neuroscience of human vision, we propose the Context-aware Two-stream Attention network (CATNet). This model makes inferences about the target object by guiding attention towards regions with informative contextual cues and object parts via dynamic integration of foveal (object) and peripheral (context) vision. The model automatically learns contextual reasoning strategies. In real world applications,  models are required to extrapolate to a wide range of different contexts, just like humans. Therefore, we test CATNet and state-of-the-art in-context object recognition models on the same psychophysics tasks \emph{without re-training the models for each experiment}. CATNet surpasses other computational models in these experiments and approximates human recognition abilities despite  the  enormous  amount  of extrapolation required.


\section{Related Works}\label{sec:relatedworks}

\subsection{Role of Context in Human Visual Recognition}
Many behavioral studies \cite{bar2003cortical,goh2004cortical} have focused on comparing congruent versus incongruent context conditions: objects appearing in a familiar background can be detected more accurately and faster than objects in an unusual scene (Fig.~\ref{fig:introincongrueg}). Several qualitative demonstrations showed that context can help visual processing  \cite{auckland2007nontarget,biederman1982scene,hollingworth1998does,aminoff2006parahippocampal}, during recognition tasks \cite{auckland2007nontarget,davenport2004scene}, detection tasks \cite{biederman1982scene,hollingworth1998does}, working memory  \cite{friedman1979framing,aminoff2006parahippocampal}, and visual search \cite{henderson1999effects}. Here we systematically tested the three fundamental properties of context to quantitatively model where, when and how contextual information modulates recognition.

\subsection{Role of Context in Computer Vision}

Contextual reasoning about objects and relations is critical to machine vision. Deep nets for object recognition, trained on natural image datasets, \eg ImageNet \cite{krizhevsky2012imagenet}, rely implicitly but strongly on context \cite{geirhos2018imagenet,brendel2019approximating}. Indeed, these algorithms often fail when objects are placed in an incongruent context (\cite{beery2018recognition,dvornik2018modeling,choi2012context}, Fig.~\ref{fig:introincongrueg}).

Many exciting successes of computer vision methods can be partly ascribed to capitalizing on the statistical correlations between contextual information and object labels. Here we briefly and non-exhaustively introduce context-aware computational models in various applications.
Qualitative analyses based on the statistical summary of object  relationships, have provided an  effective source of information for perceptual inference tasks, such as object detection (\cite{torralba2003contextual,park2010multiresolution, hoiem2005geometric, torralba2010using,liu2018structure}), scene classification (\cite{gonfaus2010harmony,torralba2005contextual,yao2012describing}), semantic segmentation (\cite{yao2012describing}), and visual question answering (\cite{teney2017graph}).

Classical approaches, \eg Conditional Random Field (CRF), reason jointly across multiple computer vision tasks in image labeling, scene classification \cite{gonfaus2010harmony,yao2012describing,ladicky2010graph,chen2018deeplab}, object detection, and semantic segmentation \cite{mottaghi2014role}. Several graph-based methods incorporating contextual information, combined with neural network architectures, have been successfully applied in object priming \cite{torralba2003contextual}, place and object recognition \cite{wu2018learning,torralba2005contextual}, object detection \cite{chen2018iterative,liu2018structure}, and visual question answering \cite{teney2017graph}. Recent interesting approaches have used deep graph neural networks for contextual inference \cite{hu2016learning,choi2012unified,deng2016structure,battaglia2016interaction}.
These works typically assume that full contextual information is always available. However, in our experiments, we include experimental conditions where partial contextual information is available, such as minimal context,  blurred context and only low-level context texture (Fig.~\ref{fig:expsintroall}). 
Breaking away from these previous works where graph optimization is performed globally, the model proposed here selects relevant visual features using an attention mechanism, and integrates partial information from both the target object and the context over multiple steps. Importantly, the model generalizes to context variations (Sec.~\ref{sec:resultsHM}). Furthermore, we provide a direct comparison against human benchmark performance.

\section{Human psychophysics experiments}\label{subsec:humanexp}
We examined the three fundamental properties of contextual modulation in recognition (Fig.~\ref{fig:expsintroall}, S1, S7, S9, S11): [A] context amount, [B] context content, [C] context dynamics. We conducted 10 psychophysics experiments, schematically illustrated in Fig.~\ref{fig:expsintroall}h, on Amazon Mechanical Turk \cite{turk2012amazon}. We recruited 80 subjects per experiment, yielding a total of $64,000$ trials (Sec.~\ref{sec:resultsHM}). \\

\vspace{-2mm}

\noindent \textbf{Experiment setup:} The stimuli consisted of 2,259 images spanning 55 object categories from the test set of the MSCOCO Dataset \cite{lin2014microsoft}.
We constrained the size of target objects to four bins : Size 1 [16-32 pixels], Size 2 [56-72], Size 4 [112-144], and Size 8 [224-288]. These bins refer to the number of pixels in the object regardless of their sizes in physical world. Given the image size of $1024 \times 1280$ pixels, and viewing distance of $\approx 0.5$ meters, these values correspond to about 1, 2, 4, and 8 degrees of visual angle (but this may vary in MTurk depending on viewing conditions).
To avoid any biases and potential memory effects, we took the following precautions. (a) Only one target object was selected per image. (b) Target objects were uniformly distributed over the 4 sizes and 55 categories. (c) Subjects saw at most 2 target objects per category. (d) The trial order was randomized.
\\

\vspace{-2mm}

\noindent \textbf{Performance evaluation and statistics:}
Most recognition experiments enforce N-way categorization (e.g., \cite{tang2018recurrent}). Here we introduced a more unbiased probing mechanism whereby there were no constraints on the words used to describe the target object  (Fig.~\ref{fig:expsintroall}h, Sec. S6.1). To evaluate human performance, we \emph{separately} collected a distribution of ground truth answers for each target object with infinite viewing time and full context (Mturk subjects \emph{not} participating in the main experiments). Answers in the main experiments were correct if they matched any of the ground truth responses.

Although computational models (Sec.~\ref{sec:model}) were evaluated using N-way categorization, we find it instructive to plot model results alongside human behavior for comparison purposes. We also show human-model correlations to describe their relative trends across conditions.
We use the Wilcoxon ranksum test  \cite{harris2013exact}, and one-way/two-way ANOVA tests \cite{ito19807} (Sec. S6.2) for statistical comparisons.

\begin{figure*}[t]
\begin{center}
\includegraphics[width=14cm]{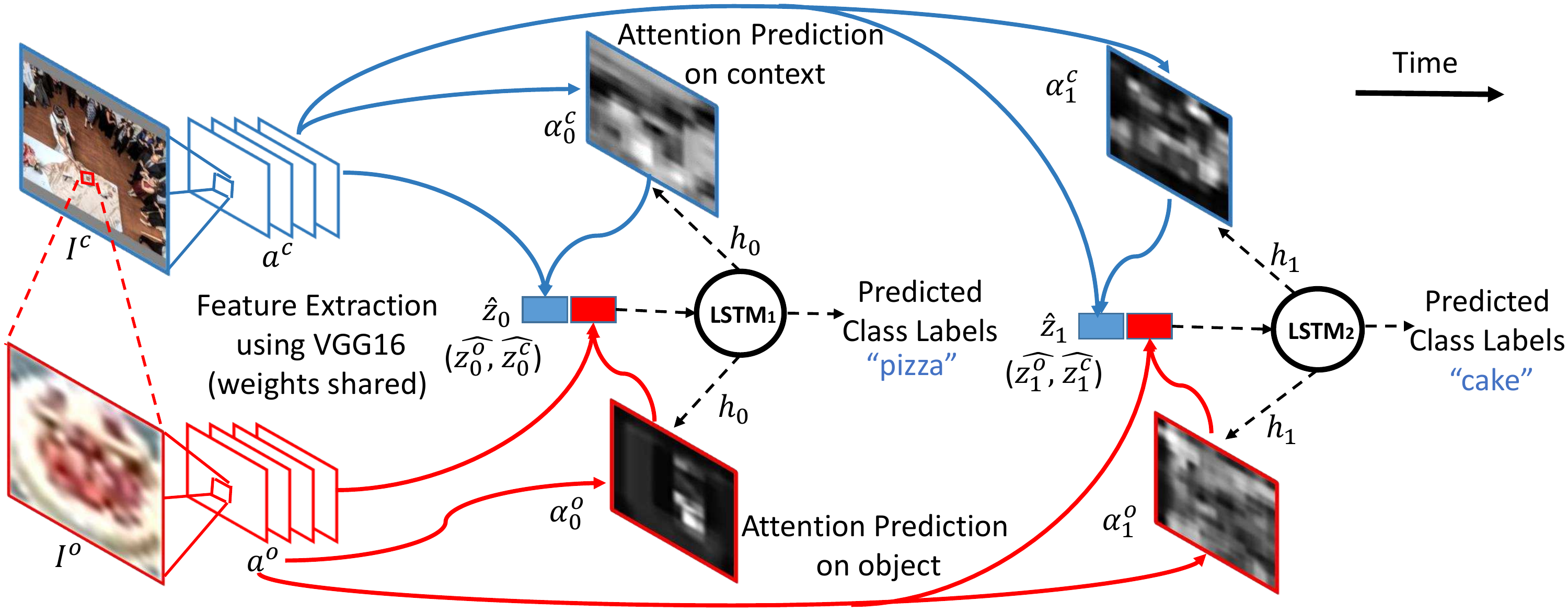}\vspace{-4mm}
\end{center}
   \caption{\textbf{Architecture overview of Context-aware Two-stream Attention network (CATNet)}. The diagram depicts the iterative modular steps carried out by CATNet over multiple time steps in the context-aware object recognition task. CATNet consists of 3 main modules: feature extraction, attention, and recurrent memory. These three modular steps repeat for a pre-specified number of time steps $T_m$. For illustrative purposes, only the first and second time steps are shown here (see Sec.~\ref{sec:model} for definition of variables and Fig. S13-S14 for implementation of the attention and LSTM modules). CATNet is \emph{only} trained using full context natural images and then it is tested in different conditions specified by each experiment (Sec.~\ref{subsec:humanexp} and~\ref{sec:resultsHM}).}\vspace{-4mm}
\label{fig:model}
\end{figure*}

\vspace{-2mm}

\section{Context-aware Two-stream Attention Net}\label{sec:model}

\vspace{-2mm}

We propose a Context-aware Two-stream Attention network (CATNet), extending work on image captioning \cite{xu2015show}.
CATNet is presented with a natural image where the target object is indicated by a white bounding box. Inspired by the eccentricity dependence of human vision, CATNet has one stream that processes only the target object ($I^o$, minimal context, Fig.~\ref{fig:expsintroall}b), and a second stream devoted to contextual information in the periphery ($I^c$, full context, Fig.~\ref{fig:expsintroall}a). The two streams are processed in parallel through weight-sharing convolutional neural networks.
$I^o$ is enlarged to be the same size as $I^c$, such that each convolutional kernel sees $I^o$ at finer-grain details.
CATNet explicitly integrates the fovea and periphery via concatenation and makes a first attempt to predict a class label $y_0$ out of a pre-defined set of $C=55$ object classes.

Horizontal and top-down connections are presumed to be important for recognition \cite{tang2018recurrent}. We added a recurrent LSTM module in CATNet to iteratively reason about context. The LSTM module changes its internal representation of the scene via attention, and predicts class labels over multiple time steps $t$ where $t \in \{1,...T_m\}$.
We use superscripts $c$ or $o$ to distinguish processes on $I^c$ or $I^o$ and subscript $t$ for time-dependent variables.

\vspace{-2mm}

\subsection{Convolutional Feature Extraction}
 CATNet takes $I^c$ and $I^o$ as inputs and uses a feed-forward convolutional neural network to extract feature maps $a^c$ and $a^o$, respectively. We use the VGG16 network \cite{simonyan2014very}, pre-trained on ImageNet \cite{deng2009imagenet} and fine-tune it at the training stage. To focus on specific parts of the image and select features at those locations, we preserve the spatial organization of features; thus, CATNet uses the output feature maps at the last convolution layer of VGG16.
 The parameters of both feed-forward feature extractor networks on $I^c$ and $I^o$ are shared. Since $I^o$ is the enlarged version of the target object region in $I^c$, this results in higher acuity and enhances sensitivity to details of the target object.
 We describe $a_c$ next but the same ideas apply to $a_o$.

A feature vector $\mathbf{a^c_{i}}$ of dimension $D$ represents the part of the image $I^c$ at location $i$, where $i=1,..,L$ and $L= W \times H$, and $W$ and $H$ are the width and height, respectively, of the feature map:

\vspace{-2mm}

\begin{equation}
a^c = \{ \mathbf{a^c_{1}},...,\mathbf{a^c_{L}} \}, \quad \mathbf{a^c_{i}} \in \mathbb{R}^D
\end{equation}

\vspace{-4mm}
\subsection{Attentional Modulation}

We use a ``soft-attention" mechanism \cite{ba2014multiple} to compute ``the context gist" $\mathbf{\widehat{z}^c_t}$ on $I_c$, and ``the object gist" $\mathbf{\widehat{z}^o_t}$ on $I_o$ (Fig. S13). There are two attention maps,
 on $I^c$ and $I^o$, respectively, where each stream has identical architectures but different weight parameters.
 We describe the context stream of attention but the same principles apply to the object attention map.
 For each location $i$ in $a^c$, the attention mechanism generates a positive scalar $\alpha^c_{ti}$, representing the relative importance of the feature vector $\mathbf{a^c_{ti}}$ in capturing the context gist. $\alpha^c_{ti}$ depends on the feature vectors $\mathbf{a^c_{i}}$, combined with the hidden state at the previous step $\mathbf{h_{t-1}}$ of a recurrent network described below:
\vspace{-2mm}
\begin{equation}\label{equ:attention}
e^c_{ti} = A^c_h \mathbf{h_{t-1}} + A^c_a \mathbf{a^c_{i}}, \quad \alpha^c_{ti} = \frac{\exp(e^c_{ti})}{\sum_{i=1}^L \exp(e^c_{ti})}
\end{equation}
\vspace{-2mm}

\noindent where $A^c_h \in \mathbb{R}^{1\times n}$ and $A^c_a \in \mathbb{R}^{1\times D}$ are weight matrices  initialized randomly and learnt during training. Because not all attended regions might be useful for context reasoning,
the soft attention module also predicts a gating vector $\beta^c_t$ from the previous hidden state $h_{t-1}$, such that $\beta^c_t$ determines how much the current observation contributes to the context vector at each location: $\beta^c_{t} = \sigma(W^c_{\beta} \mathbf{h_{t-1}})$, where $W^c_{\beta} \in \mathbb{R}^{L\times n}$ is a weight matrix and each element $\beta^c_{ti}$ in $\beta^c_t$ is a gating scalar at location $i$. $\beta^c_t$ helps put more emphasis on the salient objects in the images ~\cite{xu2015show}. Once the attention map $\alpha^c_t$ and the gating scale $\beta^c_t$ are computed, the model applies the ``soft-attention" mechanism to compute  $\mathbf{\widehat{z}^c_t}$ by summing over all the $L$ regions in the image:
\vspace{-4mm}
\begin{equation}\label{equ:contextvec}
\mathbf{\widehat{z}^c_t} = \sum_{i=1}^L \beta^c_{ti} \alpha^c_{ti} \mathbf{a^c_{i}}
\end{equation}
\vspace{-2mm}

We define $\mathbf{\widehat{z}_t} = (\mathbf{\widehat{z}^c_t},\mathbf{\widehat{z}^o_t})$ as concatenation of $\mathbf{\widehat{z}^c_t}$ and $\mathbf{\widehat{z}^o_t}$, which is used as input to an LSTM module.
The attention module is smooth and differentiable, and CATNet learns all the weights end-to-end via back-propagation.


\subsection{Recurrent Connections using LSTM}\label{subsec:lstmmod}
\vspace{-2mm}

A long short-term memory (LSTM) network predicts the class label $y_t$ based on the previous hidden state $\mathbf{h_{t-1}}$ and the gist vector $\mathbf{\widehat{z}_t}$ for $I^o$ and $I^c$  \cite{zaremba2014recurrent} (Fig. S14).
The variables $\mathbf{i_t}, \mathbf{f_t}, \mathbf{c_t}, \mathbf{o_t}, \mathbf{h_t}$ represent the input, forget, memory, output and hidden state of the LSTM (Section S3).
To compare CATNet and human performance for different exposure time $T$ (\textbf{Exp. C}), we set one LSTM time step to be $25$ ms, and considered the CATNet predicted labels at the corresponding number of time steps $T_m = T / 25$.

To predict the class label $y_t$ for the target object, the LSTM computes a classification vector where each entry denotes a class probability given the hidden state $h_t$:
\vspace{-2mm}
\begin{equation}\label{equ:classpred}
y_t = \arg \max_c p(y_c), \quad p(y_c) \propto L_h \mathbf{\mathbf{h_t}}
\end{equation}
where $L_h \in \mathbb{R}^{C \times n}$ is a matrix of learnt parameters initialized randomly. We discuss alternative convolutional LSTM connections in Section S3. 

\subsection{Training and Implementation Details}\label{subsec:varaitionsclicknets}
\vspace{-2mm}

We trained CATNet end-to-end by minimizing the cross entropy loss between the predicted label $y_t$ at each time step $t$ and the ground truth label $x$: $LOSS = \sum_{t=1}^{T_m} (-\log(P(y_t|x)))$. Predicting labels at every time step  allows  us  to  assess  the  effect  of  image  exposure time in \textbf{Exp C} (Fig.~\ref{fig:expsintroall}h, S7, S9, S11 and  Sec.~\ref{subsec:expC}). Besides, using ground truth labels at every time step empirically helped CATNet converge faster during training.


We used all MSCOCO training set images for training and validation. On every image, each object was selected as the target, always shown in full context. Only at the testing stage, we varied the context based on different experimental conditions. Importantly, \emph{none of the human behavioral experiments were used to train the model}. Both $I^c$ and $I^o$ were $400 \times 400$ pixels. We set the total number of time steps $T_m=8$ for training CATNet. Further implementation details are provided in Sections S3 and S4.

\subsection{Competitive baselines and ablated models}
\vspace{-2mm}

We compared the results of CATNet against several competitive baselines, such as DeepLab-CRF \cite{chen2017deeplab} in semantic segmentation and YOLO3 \cite{redmon2016you,yolov3} in object detection. These models were adapted to
the context-aware object recognition task (Sec. S5).

 To study the role of attention, the two-stream architecture, and recurrent connections, we introduced ablated versions of CATNet (Sec. S5). Starting from original VGG16 object recognition network \cite{simonyan2014very} pre-trained on ImageNet \cite{deng2009imagenet}, we added in one component at a time and evaluated their incremental performance change. These models include VGG16 + binary mask, two-stream VGG16, VGG16 + attention, and VGG16 + attention + LSTM.

\section{Results}\label{sec:resultsHM}
\subsection{Exp A: Amount of context}


\subsubsection{Object size matters (Exp A1)}\label{contextsizematters}
\vspace{-2mm}

We conjectured that the  impact of contextual information would depend on the target object size. We considered 4 object sizes (Sec.~\ref{subsec:humanexp}). For each size, we introduced either minimal context (rectangular bounding box enclosing the object, Fig.~\ref{fig:expsintroall}b), or full context (entire image, Fig.~\ref{fig:expsintroall}a).

\begin{figure}
\begin{center}
\includegraphics[width=8cm]{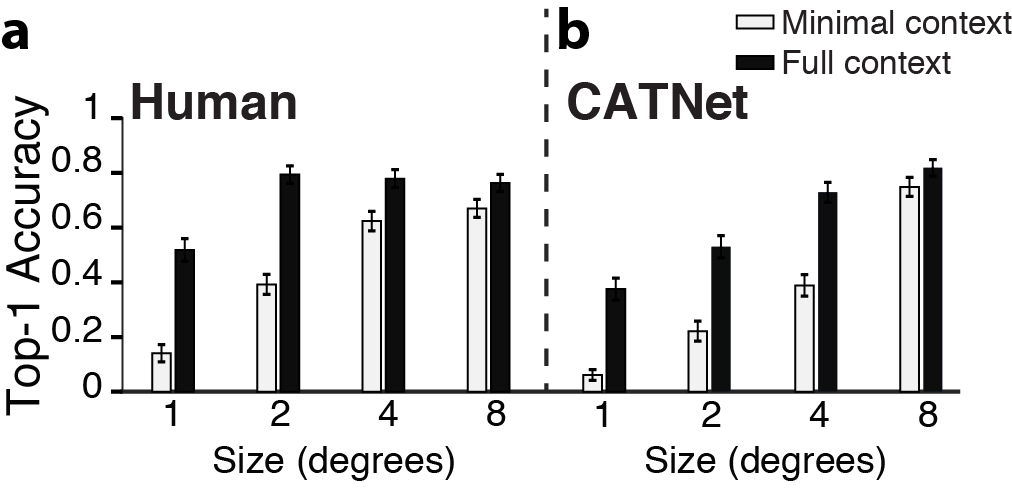}\vspace{-4mm}
\end{center}
        \caption{\textbf{Context improves recognition, particularly for small objects (Exp A1)}. Top-1 accuracy increases with object size (Fig.~\ref{fig:expsintroall}a-b). Contextual information facilitates recognition, particularly for small target objects, for humans (\textbf{a}) and CATNet (\textbf{b}). Here and in subsequent figures, error bars denote SEM. In \textbf{b}, chance level is 1/55 (see text for details).
        \vspace{-4mm}}
        \label{fig:expa1result}
        \end{figure}

For the minimal context condition (Fig.~\ref{fig:expsintroall}b), human performance improved monotonically as a function of object size from $0.140 \pm 0.031$ to $0.670 \pm 0.035$
(\textbf{Exp A1}, Fig.~\ref{fig:expa1result}, one-way ANOVA: $F(3, 5097) = 215$, $p < 10^{-15}$). This effect was readily captured by the CATNet model (one-way ANOVA: $F(3, 4368) = 304$, $p < {10^{-15}}$).

Adding full contextual information (Fig.~\ref{fig:expsintroall}a) led to a large improvement in performance both for humans and CATNet. The enhanced performance due to contextual modulation strongly depended on object size: the performance ratio between the full context and minimal context conditions was 4.7 and 2.5 (humans and CATNet, respectively) for object size 1, whereas the ratio was 1.1 and 1.05 (humans and CATNet, respectively) for object size 8. Contextual information greatly facilitates performance when the target objects are small and hard to recognize.

\vspace{-2mm}
\subsubsection{The amount of context matters (Exp A2)}\label{contextsizematters}
\vspace{-2mm}

For each object size, we systematically titrated the amount of contextual information (Fig.~\ref{fig:expsintroall}c). The context-object ratio (CO) is the total image area \emph{excluding} the target object divided by the object size. We included CO=0 (no pixels surrounding the object), 2, 4, 8, 16, and 128. Some combinations of large object sizes and large CO values were not possible.

We quantified how the amount of contextual information impacts recognition by titrating the context object ratio (CO) from 0 to 128 (\textbf{Exp A2}, Fig. S2). The amount of context was important both for humans (one-way ANOVA: $F(7, 5097) = 31$, $p < 10^{-15}$), and for CATNet (one-way ANOVA: $F(7, 4368) = 23$, $p<10^{-15}$).
Across all the CO ratios, humans outperformed CATNet for small object sizes, and CATNet outperformed humans for the largest object size. Of note, CATNet was \textit{never} trained or fine-tuned with the human psychophysics measurements. These experiments demonstrate that the context \emph{quantity} can strongly enhance recognition; which we refer to as \emph{context modulation} for short throughout the rest of the text.

\subsection{Exp B: Context Content}

\noindent We studied how context resolution, geometry, and congruency  modulate recognition in 5 experiments, focusing on object sizes 1/2/4, with minimal/full context.

\vspace{-2mm}

\subsubsection{Blurred context is sufficient (Exp B1)}
\vspace{-2mm}


\begin{figure}
\begin{center}
\includegraphics[width=8cm]{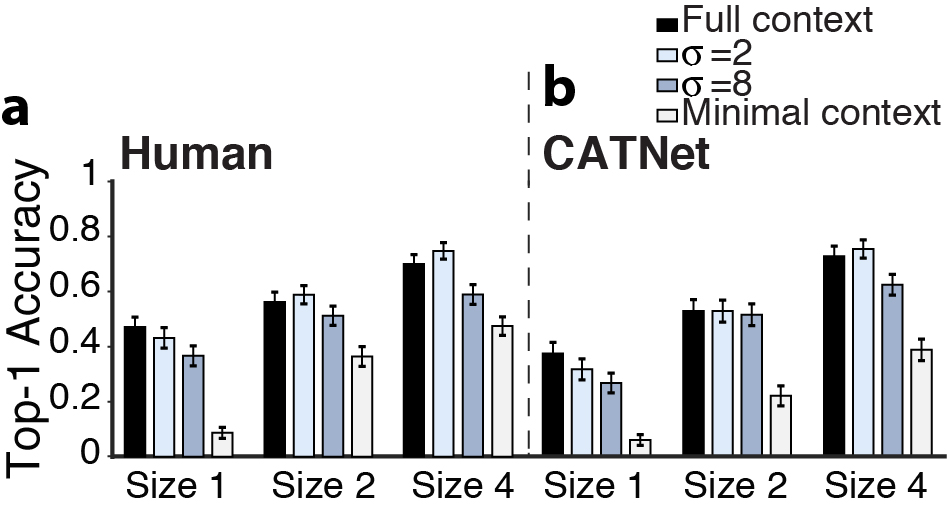}\vspace{-4mm}
\end{center}
        \caption{\textbf{Contextual facilitation persists even after small amounts of blurring (Exp B1)}. A large amount of context blurring (Fig.~\ref{fig:expsintroall}d) is required to disrupt contextual facilitation for humans (\textbf{a}) and CATNet (\textbf{b}).  Only $\sigma=2$ and $\sigma=8$ are shown here (see Fig. S3 for intermediate $\sigma$ values). 
        }
        \vspace{-4mm}
        \label{fig:expb1result}
\end{figure}

Due to the strong eccentricity dependence of human vision, the periphery has less resolution than the fovea (resolution drops so sharply that humans are legally blind in the far periphery). We conjectured that low resolution context could be sufficient to facilitate recognition. To quantify the impact of context resolution, we blurred the context  (Fig.~\ref{fig:expsintroall}d) using a zero-mean Gaussian with standard deviation $\sigma=2, 4, 8, 16, 32$ pixels (image size = $1024\times1280$ pixels) (\textbf{Exp B1}, Fig.~\ref{fig:expsintroall}d). Each subject saw all blurred conditions, with different images.


Accuracy dropped from levels indistinguishable from the full resolution condition when $\sigma \le 8$ pixels to the minimal context condition levels when $\sigma = 32$ pixels (Fig.~\ref{fig:expb1result}, one-way ANOVA: $F(4, 2933) = 28$, $p < 10^{-15}$, Fig. S3).
Interestingly, there was a wide range of blurring that led to robust context modulation, consistent with the notion that humans do not require full resolution context. The effects of blurring were also captured by CATNet, where contextual modulation disappeared only when using large $\sigma$ values (one-way ANOVA: $F(4, 2354) = 2$, $p < 0.05$). Similar to  \textbf{Exp A1-A2}, humans outperformed CATNet on small objects.


We compared the effect of blurring the context versus the target object by applying the same Gaussian blurring (\textbf{Exp. B1}) only to the object itself (\textbf{Exp. B2}, Fig. S4).
Although the number of pixels affected by blurring the target object was much smaller than blurring the context (for a fixed $\sigma$), modifying the object led to larger accuracy drops, for object sizes 2 and 4, both for humans and CATNet.



\begin{figure}
\begin{center}
\includegraphics[width=8cm]{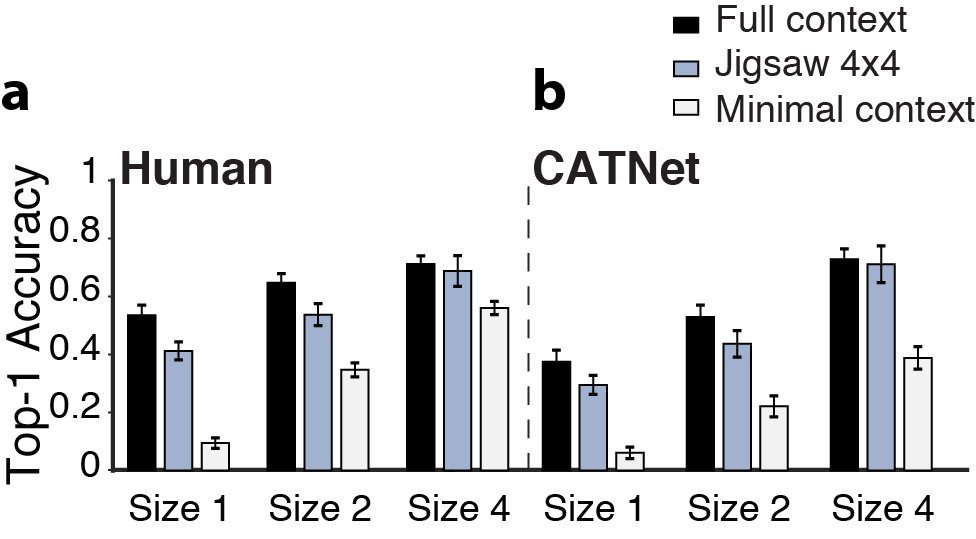}\vspace{-4mm}
\end{center}
\caption{\textbf{Large geometrical context re-arrangements disrupt contextual facilitation (Exp B4)}. Scrambling context pieces (Fig.~\ref{fig:expsintroall}f) reduced facilitation only when many small context pieces were changed, both for humans (\textbf{a}) and CATNet (\textbf{b}). Only the 4x4 condition is shown here (see Fig. S6 for other conditions).
\vspace{-3mm}
  }\label{fig:expb4result}
\vspace{-4mm}
\end{figure}

\vspace{-3mm}
\subsubsection{Contextual effects rely on spatial configuration}
\vspace{-2mm}


The relative position of objects and features in an image also affects recognition; \eg, the sky is often at the top. To evaluate the impact of contextual configuration, we randomly scrambled the images into $2 \times 2$, $4 \times 4$, or $8 \times 8$ "jigsaw" pieces while the piece with the target object remained in the same position (\textbf{Exp B4}, Fig.~\ref{fig:expsintroall}f).
We did not consider cases when the object occupied more than one piece. Both humans and CATNet relied on the spatial configuration of context over all object sizes (Fig.~\ref{fig:expb4result}, humans: one-way ANOVA: $F(3, 2182) = 58$, $p < 10^{-15}$; CATNet: one-way ANOVA: $F(3, 1787) = 29$, $p < 10^{-15}$, Fig. S6). The inconsistent spatial configuration of contextual information in the $4 \times 4$ and $8 \times 8$ configurations led to a reduction in accuracy. Interestingly, the $2 \times 2$ configuration was not different from the unscrambled full context condition, probably because each large piece already contained sufficient contextual information, and context reasoning decreases with distance to the target~\cite{zhang2019lift}.

CATNet was more robust to the distorted spatial configurations: recognition accuracy differed from the full-context condition only for the $8 \times 8$ configuration (for $2 \times 2$ and $4 \times 4$, two-tailed ranksum test, $p\geq0.12$).

\subsubsection{Low-level contextual properties do not lead to facilitation (Exp B3)}
\vspace{-2mm}


Given that the moderately blurred context still facilitated recognition (Fig.~\ref{fig:expb1result}), we asked whether low-level texture features could also enhance performance. We constructed textures constrained by the image statistics \cite{portilla2000parametric}, and pasted the intact object on them in their original locations (\textbf{Exp B3}, Fig.~\ref{fig:expsintroall}e). The textures preserve low-level features, but distort high-level features and semantic information.

Low-level texture features did not facilitate object recognition for either humans or CATNet (Fig. S5). In fact, human performance was actually slightly impaired when objects were embedded within these textures compared to the minimal context condition (two-tailed ranksum test, all object sizes, $p < 0.04$).
For CATNet, low-level texture features improved recognition with respect to minimal context only for object size 1, but the effect was much smaller than when using full contextual information.

\subsubsection{Incongruent context impairs recognition}
\vspace{-2mm}


Given that low-level textures did not help (and could even hurt recognition), and inspired by Fig.~\ref{fig:introincongrueg} and related experiments, we next studied recognition when objects were removed from their original images and placed in the same location but in different images with either a congruent context (object and context belong to the same class label) or incongruent context (context taken from a different image class label, Fig.~\ref{fig:expsintroall}g).

\begin{figure}[t]
\begin{center}
  \includegraphics[width=8cm]{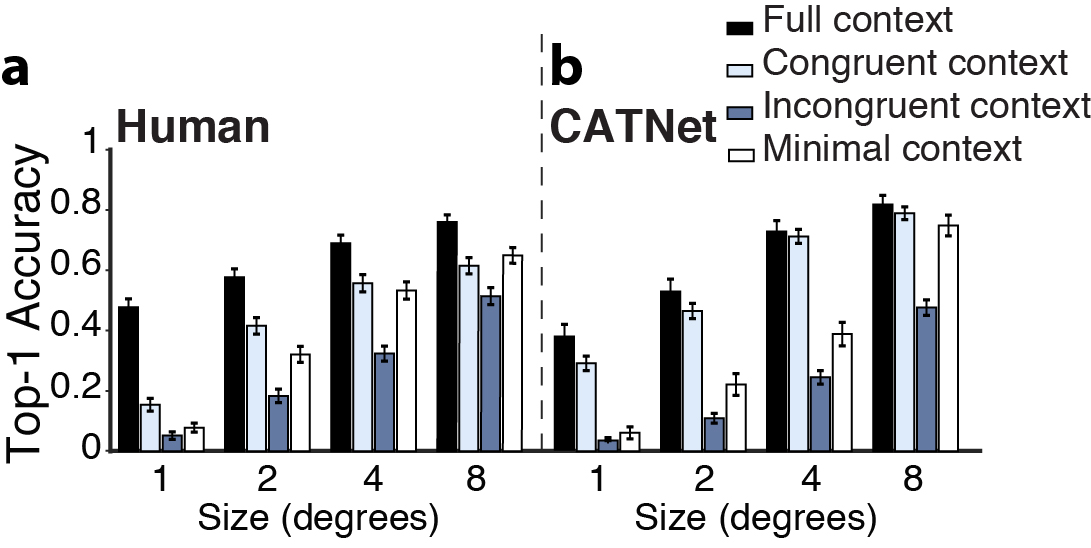}\vspace{-4mm}
  \end{center}
  \caption{\textbf{Incongruent context impairs recognition.} Pasting the target objects in different but congruent contexts facilitates recognition. Pasting the target objects in incongruent contexts  (Fig.~\ref{fig:expsintroall}g) impairs recognition, both for humans and CATNet. \vspace{-4mm} }\label{fig:expb5result}
\end{figure}

Congruent context enhanced recognition for small object sizes compared to the minimal context condition both for humans and CATNet (Fig.~\ref{fig:expb5result}). Although congruent context typically share similar correlations between objects and scene properties, pasting the object in a congruent context led to weaker enhancement. This lower contextual facilitation may be due to the erroneous relative size between objects, unnatural boundaries created by pasting, or contextual cues specific to each image. CATNet was relatively oblivious to these effects and performance in the congruent condition was closer to that in the original full context condition.

In stark contrast, incongruent context consistently degraded recognition performance below the minimal context condition. Across all object sizes, subjects showed higher accuracy for objects in congruent versus incongruent contexts (one-way ANOVA: $F(1, 2530) = 92$, $p < {10^{-15}}$).  Accuracy was lower for incongruent context than minimal context (two-tailed ranksum test, $p = 0.0005$). Similarly, CATNet recognition accuracy  also positively correlated with congruent context (one-way ANOVA: $F(1, 2977) = 515$, $p < {10^{-15}}$), and was degraded by incongruent context (for all object sizes, two-tailed ranksum test, $p < 0.001$).

\subsection{Exp C: Dynamics of Contextual Modulation}\label{subsec:expC}
\noindent The previous sections characterized \emph{spatial} aspects of contextual modulation.
The temporal dynamics of recognition places strong constraints to interpret the flow of bottom-up and top-down visual processes \cite{thorpe1996speed,tang2018recurrent,riesenhuber1999hierarchical}. Next, we conducted 3 experiments to investigate the dynamics of contextual effects on recognition.

\vspace{-2mm}

\subsubsection{Contextual modulation is fast (Exp C1)}
\vspace{-2mm}


In experiments \textbf{A} and \textbf{B}, the image duration $T$ was 200 ms (Fig.~\ref{fig:expsintroall}h). Here we systematically varied $T$ to be 50, 100, or 200 ms (Fig. S7, \textbf{Exp. C1}).
Human performance was largely unaffected by the image duration (Fig. S8). To assess the role of exposure time in CATNet, each computational time step was mapped to 25 ms (Sec~\ref{subsec:lstmmod}).
Consistent with human behavior results, exposure time had no effect on recognition for CATNet.

\subsubsection{Backward masking weakens contextual modulation (Exp C2)}
\vspace{-2mm}


The rapid computations in \textbf{Exp C1} are thought to involve largely bottom-up processing \cite{serre2007quantitative,dicarlo2012does}. Despite the short exposure, additional computations could take place after stimulus offset. The next experiment sought to interrupt those computations using backward masking  (\textbf{Exp C2}, Fig. S9).
Backward masking is commonly used in neuroscience to interrupt visual processing \cite{tang2018recurrent}. The mask shown after stimulus offset is purported to block top-down and recurrent computations. We used Portilla masks \cite{portilla2000parametric} as in \textbf{Exp B3} (Fig. S9) and stimulus exposure times followed \textbf{Exp C1}.


Backward masking did not change accuracy in the minimal context condition (Fig. S10). The recognition enhancement in the full context condition was impaired when the mask was introduced after 50-100 ms exposure, but not with 200 ms, consistent with previous studies \cite{tang2018recurrent}. In sum, contextual modulation is fast and is likely to involve recurrent computations that can be interrupted by masking. 

\subsubsection{Brief exposure to context is sufficient for facilitation (Exp C3)}
\vspace{-2mm}


In all the experiments above, object and context information were presented synchronously. During natural vision, subjects move their eyes from a given location P1 to another location P2. The information gathered while fixating at P1 acts as a prior temporal context of fixation at P2.
To investigate the effect of such prior temporal context in recognition, while conceptually simplifying the problem, we split the image into context-only and object-only parts. First, the context-only part was presented for a duration of $T1$ = 25, 50, 100, or 200 ms. Next, the context was removed, and the object-only part was presented for a duration $T2$ = 50, 100, or 200 ms (\textbf{Exp. C3}, Fig. S11). The corresponding synchronous conditions were also included for comparison purposes.


Surprisingly, even 25 ms exposure to context was sufficient to trigger contextual modulation (Fig. S12). For small objects, contextual facilitation was larger with increased context exposure, reaching the levels of the synchronous condition for 100 ms context exposure. In sum, a previous saccade,  which typically last 200 ms, provides enough contextual information that can be held in memory and enhance recognition of a minimal context object. Even shorter exposure to context already enhances recognition.

\subsection{Comparison with other models}
We have focused on the CATNet model introduced in Fig.~\ref{fig:model}. Several other computational models  incorporate some form of contextual information (Sec.~\ref{sec:relatedworks}).
We compared CATNet versus two state-of-the-art models incorporating contextual information for semantic segmentation: (1) Deeplab \cite{chen2018deeplab}), and (2) object detection (YOLO3,  \cite{yolov3}). The average accuracy across all conditions in all the variations of Experiments A and B are shown in Table~\ref{tab:avgaccu}. Details about performance of these models are shown in Fig. S16-S22 (Deeplab), and Fig. S23-S29 (YOLO3).

Although Deeplab and YOLO3 leverage on global context information, CATNet outperformed both models, especially on small objects. For example, Deeplab performed almost as well as CATNet on large objects, but it failed to demonstrate the strong contextual facilitation repeatedly observed in every experiment (Fig.~\ref{fig:expa1result},~\ref{fig:expb1result},~\ref{fig:expb4result},~\ref{fig:expb5result}).

Similarly, even though YOLO3 has a dedicated recognition module after region proposal, it failed to incorporate contextual information when recognizing small objects. We also emphasize again that \emph{all} computational models, including CATNet, performed worse than humans on small objects in every experiment, suggesting that it is necessary to come up with better ways of reasoning about context in computer vision tasks.

At least partly, the baseline models struggle with small objects due to lack of scale tolerance.
In addition to absolute accuracy, we also report  the correlation between each model and human performance across conditions for each experiment in Table~\ref{tab:avgaccu}. The correlation between the algorithms and humans reflects how each model is affected by different conditions. Not only did the baselines show lower accuracy, but they also showed lower correlation with human performance than CATNet.

\begin{table}[t]
\vspace{-6mm}
\footnotesize
\begin{tabular}{|c|c|c|c|c|c|c|c|}
\hline
  \textbf{Accuracy}               & A1   & A2   & B1   & B2   & B3   & B4   & B5   \\ \hline
Humans           & \textbf{0.58} & \textbf{0.58} & \textbf{0.47} & \textbf{0.49} & \textbf{0.39} & \textbf{0.48} & 0.43 \\ \hline
CATNet           & 0.48 & 0.48 & 0.42 & 0.39 & 0.34 & 0.41 & \textbf{0.44} \\
\scriptsize{DeepLab~\cite{chen2018deeplab}}            & 0.52 & 0.45 & 0.37 & 0.42 & 0.31 & 0.38 & 0.39 \\
\scriptsize{YOLO3~\cite{yolov3}}            & 0.26 & 0.25 & 0.13 & 0.14 & 0.13 & 0.13 & 0.19 \\ \hline
\textbf{Correlation}               & A1   & A2   & B1   & B2   & B3   & B4   & B5   \\ \hline
CATNet           & 0.89 & \textbf{0.89} & \textbf{0.95} & 0.87 & 0.89 & \textbf{0.92} & \textbf{0.93} \\
\scriptsize{DeepLab~\cite{chen2018deeplab}}            & \textbf{0.90} & 0.83 & 0.86 & \textbf{0.88} & \textbf{0.90} & 0.81 & 0.91 \\
\scriptsize{YOLO3~\cite{yolov3}}            & 0.75 & 0.78 & 0.74 & 0.78 & 0.75 & 0.66 & 0.87 \\ \hline
\end{tabular}
\caption{\textbf{Performance and correlations between humans and models for Experiments A and B}. See Sec.~\ref{subsec:humanexp} for definitions of evaluation metrics. Best is in bold. See Table S1 for additional comparisons. \vspace{-4mm}}\label{tab:avgaccu}
\end{table}

\subsection{Ablation reveals critical model components}

To distill how different components of CATNet contribute to integrating contextual information, we considered modified versions with ablated components (Tab. S1). We first evaluated pre-trained VGG16 \cite{simonyan2014very}. The accuracy of VGG16 was essentially at chance, particularly for small objects (Fig. S30-S36), confirming that in-context object recognition is not a trivial visual feature mapping task and requires focusing on the target object location.
Next, we concatenated the natural stimulus with a binary mask indicating the target object location (VGG16+binarymask). Although the binary mask increased performance with respect to VGG16, accuracy was still well below CATNet (Figs. S37-S43), suggesting that the attentional mechanism to weigh the different features plays an important role.
We therefore implemented an attention module (Sec.~\ref{sec:model}, VGG16+attention), which led to a large performance boost (Figs. S44-S50), consistent with previous work showing the efficiency of attention in vision tasks \cite{linsley2018learning}. In Fig. S15,
we provide visualization examples of predicted attention maps on context and target objects respectively. CATNet learns to focus on informative context regions for recognition. Consistent with previous work \cite{linsley2018learning}, attention on target objects is  sparse and focuses on object edges or the minimal context regions surrounding the target rather than on visual features on the targets themselves.
Additional improvement in performance was achieved by incorporating a two-stream module (Figs. S51-S57), and an LSTM module (Figs. S58-S64).

\section{Discussion}\label{sec:discussion}

We quantitatively studied the role of context in visual recognition in human observers and computational models in  tasks requiring identification of target objects in natural settings.
We investigated three critical properties of context: quantity, quality, and dynamics. Contextual facilitatory effects were particularly pronounced for small objects and increased with the amount of peripheral information. The notion of full context used here and in most computer vision databases is arbitrarily defined by the person taking the picture (as opposed to a true full image which could be defined by the entire human visual field). Thus, Experiment A provides a direct titration of how different amounts of context impact recognition (Fig.~\ref{fig:expa1result}, Fig. S2).

Consistent with the eccentricity dependence of human vision, facilitation was not affected by small amounts of blurring (Fig.~\ref{fig:expb1result}, Fig. S3), or geometrical rearrangements that left intact information near the target object (Fig.~\ref{fig:expb4result}). Congruent contextual information typically enhanced recognition, while incongruent context impaired performance (Fig.~\ref{fig:expb5result}). Contextual effects could not be accounted for by low-level image properties (Fig. S5). Such contextual modulation happened fast (Fig. S7-S8), and could even be elicited in an asynchronous fashion where the context was shown before the target object (Fig. S11-S12). Contextual modulation was impaired by rapid interruption via backward masking (Fig. S9-S10).


To compare against the benchmark of human-level in-context recognition, we evaluated competitive methods in computer vision, and introduced a recurrent neural network model (CATNet, Fig.~\ref{fig:model}). CATNet combines a feed-forward visual stream module that dynamically extracts image features with an attention module to prioritize different image locations. CATNet integrates information over time,  producing a label for the target object. Surprisingly, even though the model lacks human expertise in interacting with objects in their context, CATNet adequately demonstrated human-like behavioral characteristics, and reached almost human-level performance in a wide range of in-context recognition tasks. However,  there are still significant gaps between models and humans, particularly when recognizing small objects within context, and also for large objects out of context.

Even though context is often only \emph{implicitly} incorporated in current algorithms, contextual information is critical to vision applications (\emph{e.g.}, object and action recognition). Dissociating the contributions of object and context helps us better interpret computer vision models. In addition, context can be used to fool current algorithms (\emph{e.g.}, Fig. 1). Thus, the experiments presented here help us understand models' failure cases. Explicitly incorporating contextual cues can further help protect computer vision models against context-based adversarial attacks. These results introduce benchmarks to integrate object recognition and scene understanding, and provide initial steps to understand human visual recognition and improve  intelligent computer vision systems.





\vspace{-2mm}

\section*{Acknowledgements}
\vspace{-2mm}
\noindent This work was supported by NIH R01EY026025 and by the Center for Minds, Brains and Machines, funded by NSF STC award CCF-1231216. MZ is supported by postdoctoral fellowship of Agency for Science, Technology and Research. We thank Martin Schrimpf for initial discussions that inspired this work, and Pranav Misra and Kasper Vinken for comments on the paper.

{\small
\bibliographystyle{ieee_fullname}
\bibliography{egbib}
}

\end{document}